\pdfoutput=1

\documentclass[11pt]{article}

\usepackage{emnlp2023}

\usepackage{times}
\usepackage{latexsym}
\usepackage{multirow}
\usepackage{pgfplots, pgfplotstable}
\pgfplotsset{compat=1.18}
\usepackage{natbib}
\usepackage[T1]{fontenc}

\usepackage[utf8]{inputenc}

\usepackage{microtype}

\usepackage[portuguese, english]{babel}
\usepackage{inconsolata}
\usepackage{color}
\usepackage{amsmath}
\usepackage{hyperref}
\usepackage{cleveref}
\usepackage[acronym]{glossaries}
\usepackage{booktabs}
\usepackage{caption} 
\usepackage{csquotes}
\usepackage{tikz}
\usepgfplotslibrary{statistics}
\newcommand*\circled[1]{\tikz[baseline=(char.base)]{
            \node[shape=circle,draw,inner sep=2pt] (char) {#1};}}
\usepackage[multiple]{footmisc}
\usepackage{bigfoot}

\DeclareNewFootnote{AAffil}[arabic]
\DeclareNewFootnote{ANote}[fnsymbol]

\newacronym{nl}{NL}{natural language}
\newacronym{ml}{ML}{Machine learning}
\newacronym{mturk}{MTurk}{Amazon Mechanical Turk}
\newacronym{ted}{TED}{Tree Edit Distance}
\newacronym{ast}{AST}{Abstract syntax tree}
\newacronym{nl2agg}{NL2Agg}{Natural Language to Aggregate}
\newacronym{nl2sql}{NL2SQL}{Natural Language to SQL}
\newacronym{ga}{GA}{General Availability}
\newacronym{cd}{CD}{Constrained Decoding}
\newacronym{cfg}{CFG}{context-free-grammar}
\newacronym{llm}{LLM}{Large Language Model}

\title{Natural language to SQL in low-code platforms}

\author{
    Sofia Aparicio\Thanks{\hspace{3pt}Work done while working at OutSystems.} \\ Tripadvisor, Portugal
    \And
    Samuel Arcadinho\FootnotemarkAAffil{*} \\ Zendesk, Portugal
    \And
    João Nadkarni \\ OutSystems, Portugal
    \AND
    David Aparício\FootnotemarkAAffil{*} \\ Zendesk, Portugal
    \And
    João Lages \\ OutSystems, Portugal
    \And
    Mariana Lourenço\FootnotemarkAAffil{*} \\ RedLight, Portugal
    \AND
    Bartłomiej Matejczyk \and Filipe Assunção\Thanks{For questions about the paper reach out to filipe.assuncao@outsystems.com} \\ OutSystems, Portugal
}

\begin{document}
\maketitle
\begin{abstract}
One of the developers' biggest challenges in low-code platforms is retrieving data from a database using SQL queries. Here, we propose a pipeline allowing developers to write \gls{nl} to retrieve data. In this study, we collect, label, and validate data covering the SQL queries most often performed by OutSystems users. We use that data to train a \gls{nl} model that generates SQL. Alongside this, we describe the entire pipeline, which comprises a feedback loop that allows us to quickly collect production data and use it to retrain our SQL generation model. Using crowd-sourcing, we collect 26k \gls{nl} and SQL pairs and obtain an additional 1k pairs from production data. Finally, we develop a UI that allows developers to input a \gls{nl} query in a prompt and receive a user-friendly representation of the resulting SQL query. We use A/B testing to compare four different models in production and observe a 240\% improvement in terms of adoption of the feature,  220\% in terms of engagement rate, and a 90\% decrease in failure rate when compared against the first model that we put into production, showcasing the effectiveness of our pipeline in continuously improving our feature.
\end{abstract}

\maketitle

\section{Introduction}

\quad 
Low-code platforms, such as Service Studio provided by OutSystems, are widely used for application and web development, specially by developers with limited engineering backgrounds~\citep{sahay2020supporting}. Data retrieval in low-code platforms is particularly important: it is one of the most searched topics~\citep{al2021empirical} and represents 50\% of development time according to our internal reports. 
In OutSystems, developers retrieve data from databases using an \emph{aggregate}, a visual representation of a SQL query. In \cref{subsec:osagg} we describe the different components of an \emph{aggregate}.

We propose a solution for developers where they express their intent in the form of \gls{nl}, which is automatically converted to an aggregate. Our approach addresses challenges in production settings like multiple data sources, extensive data model schemes, high throughput requirements, and UI design. Previous studies have not applied this approach to production settings~\citep{arcadinho2022t5ql, scholak2021picard, wang2019rat, yu2019cosql, yu2018spider}. We put forward a production pipeline for the task of the \gls{nl2sql} task. We explore different models for SQL generation, and use a data-driven approach to collect the right labels for supervised classification according to production data distribution. We also design a UI (see Figure~\ref{fig:aggregate} for an example) for developers to write the NL and receive the corresponding SQL query as an aggregate.



We conduct offline experiments for the \gls{nl2sql} task and achieve a good balance between accuracy and response time using a curated in-house test set that mimics our users' usage distribution. 
Additionally, we perform online A/B testing to compare model performance. 
Here, the model is evaluated according to adoption, engagement, and failure rate. 
Our SQL generation model, based on T5 with constrained decoding (as described in \citet{arcadinho2022t5ql}) increases engagement by 220\%, adoption by 240\%, and decreases the failure rate by 90\% when compared against a baseline model that we initially deployed, based on \cite{wang2019rat}.

Thus, our main contribution is a data retrieval pipeline for low-code developers with no SQL background. Our pipeline performs data collection, SQL generation, and model retraining. As far as we know, no previous study showcased a complete pipeline. We also produce an efficient model for \gls{nl2sql}: it has high accuracy, can handle large data model schemas, and rarely generates invalid SQL. Finally, we describe a UI that given \gls{nl} queries provides the users with SQL queries in a simplified visual representation (an aggregate).


 The remainder of the paper is organized as follows. Section~\ref{sec:method} describes the pipeline and its different components. Section~\ref{sec:experiments} details the experiments. We discuss related work in Section~\ref{sec:related_work} and present our conclusions in Section~\ref{sec:conclusions}.

\section{Method}\label{sec:method}

\begin{figure*}[ht]
    \centering
    \includegraphics[width=.95\linewidth]{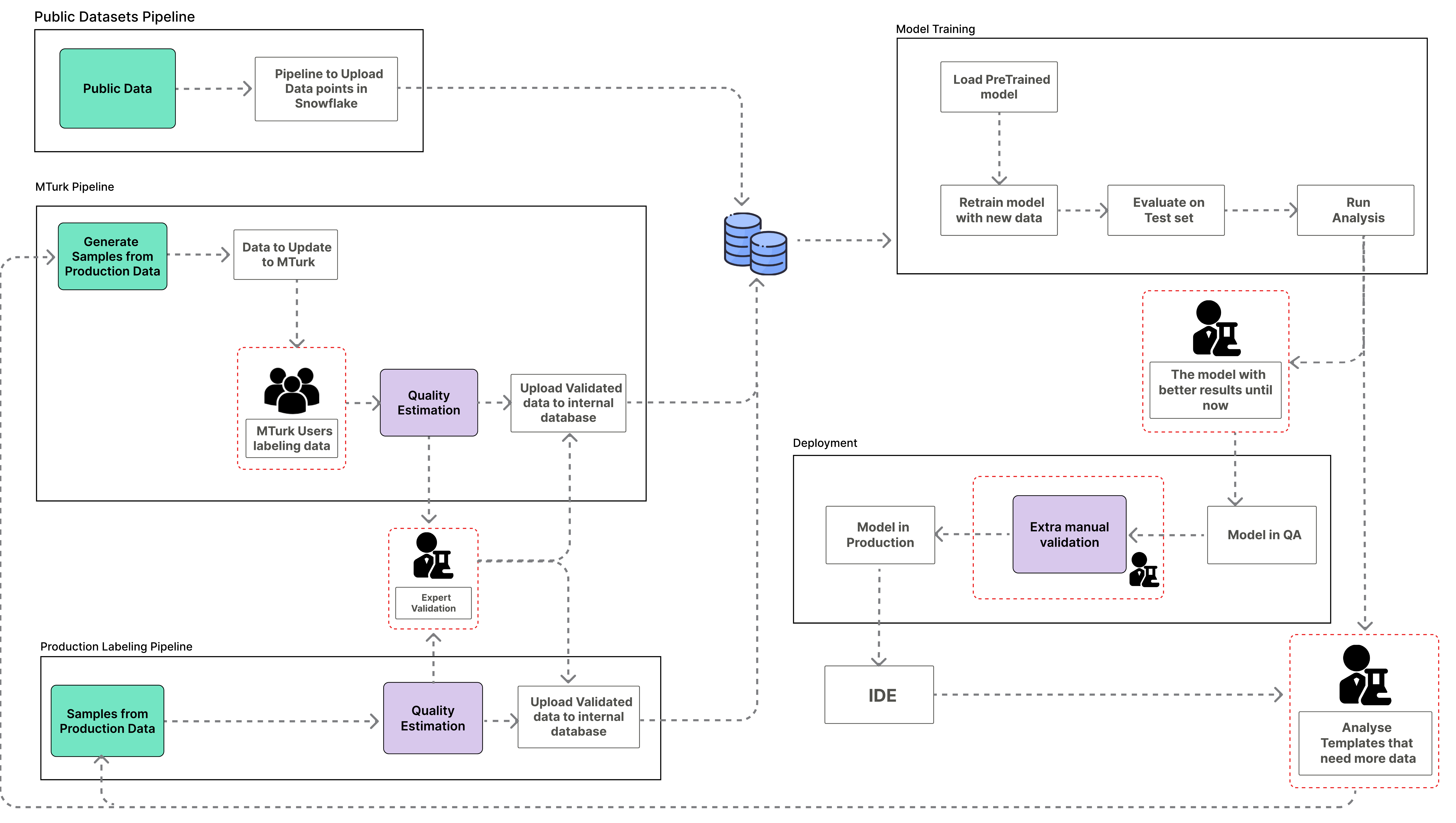}
    \caption{Machine Learning pipeline that comprises the three main components, data collections (on the left), model training (top right corner), and model deployment (bottom right corner).}
    \label{fig:pipeline}
\end{figure*}

We develop a pipeline to produce a visual representation of SQL (i.e., an aggregate) from pairs of textual utterances and data model schemas. Our pipeline has three main sections: data collection, model training, and model deployment (Figure~\ref{fig:pipeline}).

\subsection{Data collection}\label{ssec:data-collection}
\quad We combine data from three sources: public datasets, datapoints collected from \Gls{mturk}, and data from production.

\subsubsection{\textbf{Public data}}
 
We use data from Spider~\cite{yu2018spider} since it is a cross-domain public dataset with a diverse set of SQL query structures of varying difficulty. Spider comprises $\sim6$k SQL queries and $\sim10$k questions across 200 databases. While useful in production and for comparing our method with other state-of-the-art approaches, we note that Spider deviates from our requirements in the following points:

\begin{enumerate}
    \item We can only rely on the tables and column names (i.e., the database schema) because the data of our users is confidential. Spider, however, contains data points where the utterance depends on the values in the database for the creation of the appropriate SQL query.

    \item The database schemas of our users' applications are larger than those found in Spider (see Figure~\ref{fig:data-histogram}). Models, like RATSQL~\cite{wang2019rat}, cannot handle our real-world database schemas due to their size. On Spider dataset, we lack that representativeness.

    \item Spider's table and column names differ significantly from those in our real-world databases and, in general, are less complex.

    \item OutSystems-specific properties (such as variables and functions, not native to SQL) need to be predicted by the model. These properties are, obviously, not found on public datasets.

    \item The most commonly asked questions in our real-world setting are not present in public datasets. Thus, if we simply evaluate models on Spider, we do not have an accurate estimation of the model performance in production.
    
\end{enumerate}

\subsubsection{\textbf{Crowdsourced labelled data}}
To mitigate the points discussed above, we complement Spider's data with additional labeled data. We have few pairs of utterances to SQL from production, but we have many SQLs created by users. Thus, we use crowdsourcing to label SQLs created by developers with the corresponding \gls{nl}, as suggested by \citet{roh2019survey}.
We use \Gls{mturk} since it allows us to select workers with technical backgrounds. We collect $26$k pairs of SQL and the corresponding \Gls{nl} utterance. Figure~\ref{fig:MTurk_experiment} shows how the \Gls{mturk} was shown to labellers.

Since the primary goal of the data collection is to mimic our users' SQL usage distribution, we need to label a representative set of examples. However, SQLs are very granular, and thus we cannot simply find the k-most frequent SQLs. Thus, we develop templates that capture the overall structure of the SQL query (e.g., "SELECT * FROM [TABLE] JOIN [TABLE] on [COL] = [COL]"). We find the most common SQLs, and sample queries from them according to the global distribution. Then, we send queries to \Gls{mturk} according to the templates' distribution.

We validate queries for model training using a semi-automatic process to identify users with bad answers. In practice, we manually validate triggers from the automatic validation process and create a denylist for future tasks.

We use the features provided by \Gls{mturk} and do feature engineering on the labeled pairs. The resulting features will be described on the appendix~\ref{subsec:mturk}.

We use a decision tree classifier to select the threshold score to use as a trigger, i.e., if the \gls{nl} has a score above this threshold, the reviewer needs to manually validate the \gls{nl}. After each crowdsourcing iteration, we also automatically analyze the workers who had a bad performance on our tasks and provided feedback to allow them to improve their labeling. If it was the second time that the worker had a bad performance, we exclude them from future tasks. From all the experiments, we collect a total of $26$k valid data points.

\subsubsection{\textbf{Data from production}}

Having deployed the trained model to production, we then start the data collection. We start by analyzing the model performance on each template. Then, we select the ones with the worst performance and collect the correspondent SQLs. To be noted that we only select SQLs that were edited by our users and the corresponding NL.  

In the end, we will have several SQLs for the same NL, where each SQL corresponds to an edition made by the user, and we need to ensure which one of the SQLs corresponds to the given NL. We run the quality estimator (described in Section~\ref{ssec:quality-estimation}) to help with the review. After that, the data is uploaded to an internal labeling tool to be labeled. Finally, we add that data to the training set and retrain the model.

    

\subsection{Quality estimation}\label{ssec:quality-estimation}

\quad As stated in the previous section, for each pair of utterances and predicted SQL query, we collect all the changes that the user did to the SQL query. We use this information to retrain the model. The interaction can be divided into four types:

\begin{enumerate}
    \item The model predicts the right SQL, and the user does not change anything.
    \item The model predicts the wrong SQL, and the user corrects it.
    \item The model predicts the right SQL. The user changes their mind and changes the SQL. 
    \item The model predicts the wrong SQL. However, the user doesn't change it.
\end{enumerate}

We select only the interactions where the user made changes to the predicted SQL to retrain the model. However, this presents challenges since the final SQL is not always correct for the utterance (case 4), and the user might change their mind during development (case 3). As such, we build a system that can filter out wrong pairs of SQL and utterances and select the correct ones. To this end, we employ two different systems, one based on rules and another based on a quality estimator.

\subsubsection{\textbf{Rule-based systems}}
This system uses the same features described in the previous section, with the only change being the weights in the decision tree.

\subsubsection{\textbf{ML-based estimator}}
 Following the work done in \cite{arcadinho2022t5ql}, we fine-tune CodeBERT \cite{2020arXiv200208155F} to detect if a SQL query matches an utterance or not. For this, we use the training dataset of our \gls{nl} to SQL model and generate possible SQL for each utterance. Then, we annotate the pairs as being true if the SQL matches the gold SQL or if the query result is the same as the gold one. Having this model fine-tuned, for each interaction, we compute the probability of each SQL to be a pair of the given utterance and select the pair with the highest score. Finally, we filter out all the pairs that have a score below a given threshold.





\subsection{SQL generation}

\quad We use as input the \gls{nl} query of the collected SQLs (see Section~\ref{ssec:data-collection}) and our goal is to generate the corresponding SQL as output. Figure~\ref{fig:t5ql} illustrates the full pipeline of our SQL generation.

T5QL~\cite{arcadinho2022t5ql} is a Seq2Seq method that converts the input sequence into the final SQL. For instance, the input of Figure~\ref{fig:t5ql} is converted to "Give me the users sorted by country | User, Account: country" and the correct SQL is "from User select * order by country". 

T5QL comprises a generator (a T5 model~\cite{raffel2020exploring}) and a ranker (a CodeBERT model~\cite{feng2020codebert}); we note, however, that T5QL is easily adaptable to use other models as either the generator or the ranker. The generator is trained to output SQLs that correspond to the input sequence. We use beam search to output several candidates, and the ranker's task is to re-rank the generator's candidates. We observe that using a re-ranker can improve the generator's performance for SQL generation in~\cite{arcadinho2022t5ql}.

As in traditional Seq2Seq frameworks, the generator iteratively generates a token using as input the input sequence and the generation up to that point. However, unlike traditional methods, in T5QL next token generation is constrained to produce only valid SQL. To guarantee this property, we build a \gls{cfg} of SQL statements and use the parser's lookahead to fetch valid continuations (i.e., tokens) to the current generation; then, we mask all invalid tokens, thus only allowing the generator to generate valid tokens. 

The generator outputs multiple candidates by using beam search. We use the ranker to learn to re-rank the candidates by giving it as input a pair of \gls{nl} and SQL and a label informing the ranker if the two correspond to each other. To train the ranker, we give it multiple pairs where only one of them is correct and the others are: eleven generations using beam search and two random SQLs from the training data. We train CodeBERT in a cross-encoding setting.

Since, in real-world scenarios, the data model schemas can be very big (because users can have many tables and columns), we use LongT5~\cite{guo2021longt5} in production as it uses sparse attention, which allows us to have bigger input sizes without the need of large GPUs.

\subsection{Model Training}\label{sec:model-training}
After collecting data, the next step is to retrain the model. This section will focus on the training part and explain the strategy used for retraining the model. In the evaluation step, we will detail the metrics used to evaluate them and their limitations.

\subsubsection{\textbf{Training}}
As explained in Section~\ref{ssec:data-collection}, the model is trained using a new version of the training dataset in each training iteration. Data points from the SQL template where the model had the worst performance are added to the train set.

By doing this, we perform active learning where the query strategy is clustering~\cite{Zhan2022ACS} on the predicted SQL template. The cluster selection is based on the previous model's performance and the user feedback, like the average number of changes it did on a prediction.

After collecting the data from the previously trained model, we fine-tune the model. 

\subsubsection{\textbf{Evaluation}}
 When evaluating a trained model, it is important to consider the metrics computed (e.g., the model's accuracy), and the metrics that compute the inference time and the overall experience observed after deploying into production. We divide the computed metrics into two types: \emph{model performance} metrics, and \emph{server performance} metrics.
 
 To evaluate model performance, we use a combination of metrics that comprise: exact match, execution match~\cite{zhong2020semantic}, and \gls{ted}. 
 
 The first two metrics are the norm when evaluating Text-to-SQL models in the Spider benchmark~\cite{yu2018spider}. However, it is hard to relate these metrics with user perception. For instance, one query that has an error in the where conditions has the same score as a completely wrong SQL.
 
The exact match and execution match are binary, which removes any concept of distance between SQL queries. With the objective of solving these problems, we used the \gls{ted} between the \gls{ast} of the gold and predicted SQLs. Using the \gls{ted} metric, we can compute the number of changes one needs to perform to transform one SQL into the other, which closely relates to the perception of error seen by the user. Furthermore, we also give different weights when computing the \gls{ted}. For example, we penalize more an error in the selected tables rather than a selected column. The weights were selected based on the number of clicks a user needs to perform to correct the span.

Looking at the server performance, we also compare the response time on the test dataset by looking at the $50\%$, $90\%$, and $99\%$ percentiles. This is done because some improvements could impact the response time. For instance, bigger models have a slower response time.
 
Finally, as seen in Figure~\ref{fig:pipeline}, all this analysis is done at the template level, where we then average using the micro and the macro averages. Doing this helps us select the next set of SQL templates we should collect to improve our model.

\subsection{Deployment}\label{ssec:deployment}

This section will refer to deployments as introducing a new service version. It should be noted that new models result in new service versions and changes in the pre/post-processing pipelines (these steps and the model forward pass constitute our model pipeline) and in the API code. We use Amazon S3 as our model registry and FastAPI as the python web framework to build the API to call our model pipeline. This API, alongside the model pipeline code, is containerized in our service image, which runs in a Kubernetes infrastructure (we use Amazon EKS to run Kubernetes on AWS).

We use a standard GitOps-based \cite{gitops2022} CI/CD approach to automate the deployments of the service. The high-level sequence of events that happen, from implementing a change in the service code to having it in production, is described below. 

\begin{enumerate}
    \item When we are ready to release a new change in the service (be it a new model, post/pre-processing approach, or bug fix), we need to create a pull request in GitHub.
    \item Creating this PR will trigger our CI pipeline to run the unit tests on top of our changes.
    \item If the tests pass, then the model is automatically placed in the Development environment, where we can test it out through an API. 
    \item Once we are at a stage where we feel comfortable with passing these new changes to the Quality Assurance (QA) environment, we merge the pull request into the main branch. 
    \item Merging the pull request into the main branch automatically triggers integration tests.
    \item If the integration tests pass, the QA environment is updated, and a new production candidate release is created.
    \item When we feel comfortable in advancing the changes that are in QA, we manually approve the production release candidate that was created, which will trigger our CD pipeline that will put these changes in Production.

\end{enumerate}

For us it's crucial that step 7 of the above list happens without downtime, meaning that we can deploy a new version of the service without making it unavailable for the users. All of our deployment strategies follow this principle. Various deployment strategies follow this, and in practice, we alternate between three different deployment strategies based on the need and release content:

\begin{itemize}
    \item \textbf{Rolling Update}: this is the simplest approach to update our service. It will simply replace the old version of the service with the new instance (Kubernetes pod) at a time. With this strategy, users are served by the new service version in a few minutes. In practice, we just use this strategy for low-risk feature bug fixes.
    \item \textbf{Progressive Rollout}: This strategy is similar to rolling update, but only updates a fraction of instances instead of all of them. For example, a progressive rollout starting at 20\% involves updating 20\% of resources with the new model version, receiving 20\% of incoming requests. The remaining resources and requests are kept for the old version of the service. Once we are confident that the new service version is responding well, we advance this progressively (looking at the service metrics, like errors and latency, for the new version before progressing), focusing on service metrics like errors and latency, until 100\% requests and resources are being used by the new version. This approach minimizes deployment risk by only progressing to more users after ensuring the service is responding as expected. This strategy is used when deploying code changes with risk, such as optimizations in pre/post-processing pipelines or service API code, without a new feature.
    \item \textbf{A/B testing}: in the A/B testing deployment strategy, we updated a fraction of the instances with the new model version, but unlike the progressive rollout strategy, this fraction doesn't increase progressively.  Instead, an experiment with this split fraction (which we usually have at 50\%) is started. After statistical significance is reached and a winning feature is found, the experiment ends. All the resources are updated (with the usual rolling update) so that the users start being only served by the winning feature. Thus, this deployment strategy is associated with starting an experiment to measure the impact (based on business metrics) of the new service version. This is the most common deployment strategy we use when deploying a new model or constraining the decoding version. We do this because we want to go beyond the offline metrics we evaluated in the lab and check if the new model can positively impact the users. Some A/B testing experiments can be found in \ref{ssec:online-experiment}.
\end{itemize}

\subsection{Feedback Loop}\label{ssec:feedbackloop}

Once our model is deployed and serves our customers, we start to have user feedback on our feature. It's crucial to gather this user feedback to set up a feedback loop where we use this user feedback data to retrain our model. This feedback loop sets up a virtuous cycle of data: the user feedback data is used to create better models, which make the product receives more usage and, respectively, more user feedback. That feedback can be used to improve the model's performance further.

There are two types of user feedback, implicit feedback and explicit feedback. Both have advantages and drawbacks and sometimes combining both can lead to the best results, as observed in \cite{feedback2018}. For this feature, the desired experience dictated that we only have user-implicit feedback. We collect user feedback through the product's telemetry data. Then, we can observe if the user maintained the suggestion provided by the model or if the suggestion was changed or deleted. When the suggestion is changed, we also collect the final state of the user's SQL query.

We correlate these telemetry events, which we can use to find our gold SQL (the final SQL query after the user's editions). With the input data, our service stores each request (the \gls{nl} query and data model schema). By merging these data sources, we get our labeled data through user implicit feedback.

\begin{table}[b!]
    \centering
    \caption{Internal dataset used in the offline experiment.}
    \label{tab:dataset_info}
    \begin{tabular}{lll}
    \cline{2-3}
                  & Train   & Test     \\ \hline
    Instances     & 29 000  & 4 860   \\
    Public         & 25.86\% & 0\%     \\
    Crowdsourcing & 73.26\% & 97.84\% \\
    Internal      & 1.88\%  & 2.16\%  \\
    \end{tabular}
\end{table}

\section{Experiments}\label{sec:experiments}

As mentioned in Section~\ref{ssec:deployment}, before releasing a new model for all users, we first evaluate its potential impact on a sample of the users. In this section, we analyze the performance of a set of models. First, we perform offline experiments on an internal test set. Then, we share the results computed in a sample of users through A/B testing. All the results are compared against a baseline, the RATSQL+GAP~\cite{shi2021learning} fine-tuned in our internal dataset. 
The composition of the internal dataset is shown in Table~\ref{tab:dataset_info}. We note that the validation dataset is a subset of the train dataset, thus we omit it.

We run our models for all the tests in a G4DN.xlarge machine with a Tesla V100 with 16GB of GPU memory. This setup allows the test to run in less than 12h.

\subsection{Offline experiments}\label{ssec:offline-experiment}
Our first deployed model was a version of RATSQL+GAP, which we refer to as \emph{baseline} henceforth. We achieved two major improvements since the first deploying this model. The first improvement was to use T5QL; \cite{arcadinho2022t5ql} shows that augmenting T5 with \gls{cd} and that adding a re-ranker of the generator's predictions can boost the performance of small models (e.g., T5-Base). The second major improvement was the introduction of context rules into \gls{cd}, which constrains even more the generation based on the context, and reducing the numerical precision of the model. The results of each improvement are summarized in Table~\ref{tab:offline-testing} and discussed next.

\subsubsection{\textbf{Maximum input size}}
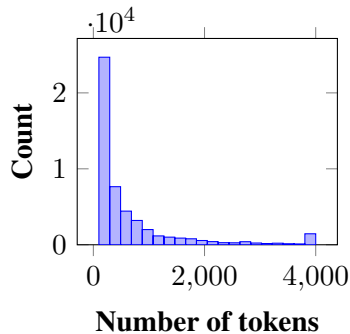
\begin{figure}[t!]
  \begin{center}
    \begin{tikzpicture}
    \begin{axis}[
        ybar,
        ymin=0,
        xlabel=\textbf{Number of tokens}, 
        ylabel=\textbf{Count},
        width=0.65\linewidth,
    ]
    \addplot +[
        hist={
            bins=20,
            data min=100,
            data max=4000
        }   
    ] table [y index=0] {data/data_model_sizes.csv};
    \end{axis}
    \end{tikzpicture}
    \caption{Input size for a sample of our data. The sample used had a size of 50000.} \label{fig:data-histogram}
  \end{center}
\end{figure}

The max input size is where we see the biggest improvement. This is justified by the fact that the RATSQL+GAP~\cite{shi2021learning} uses a fine-tuned version of the BART encoder~\cite{lewis2019bart}, which has a maximum input size of 1024 tokens. However, our token length distribution shows that the real data has instances with more tokens than what the baseline model supports (see Figure~\ref{fig:data-histogram}).

Using the settings described previously, we increase the input size from 1024 tokens to 20k tokens. This limit was further pushed to 26k by allowing the model to run on FP16 precision.

\subsubsection{\textbf{Response time}}
The response time measures each model's average time to answer a request. In Table~\ref{tab:offline-testing}, one can see that the first iteration of the T5QL model increased the response time of our model, however, by reducing the precision of our model we achieve competitive results with the baseline.

\subsubsection{\textbf{Errors}}
Finally, analyzing the TED of each improvement, one can observe that by reducing the numerical precision of the model, the TED also increases. This is expected since FP16 introduces small rounding errors that affect the final prediction. Nevertheless, the final error rate is still significantly better than our defined baseline by $\approx654\%$.






\begin{table}[b!]
\centering
\caption{Deployed models performance. The baseline is RATSQL+GAP~\cite{wang2019rat}, T5QL is the model proposed in \cite{arcadinho2022t5ql} and CD+FP16 is the same as \cite{arcadinho2022t5ql} but with added business rules and lowered precision.}
\label{tab:offline-testing}
\begin{tabular}[b]{lllll}
\cline{2-4}
            & Max input & Response & TED  \\ 
            & size & Time (s) &   \\ \hline
Baseline    & 1 024          & 2.12              & 6.93 \\
T5QL        & 20 000         & 10.37             & 0.89 \\
T5QL + FP16 & 26 000         & 2.38              & 1.06 \\ \hline
\end{tabular}
\end{table}

\subsection{Online experiments}\label{ssec:online-experiment}

After the deployment of the model to production and during the A/B testing phase, we have a dashboard to track the experiment and monitoring systems to detect failures. We measure adoption, engagement, and success during the experiments. In the following section, we explain each metric and detail the results in the following subsections. 

\subsubsection{\textbf{Adoption}} Measured by two different metrics: the adoption rate and the adoption funnel. The adoption rate is the percentage of users who have used the \gls{nl2agg} feature, the percentage of users that created aggregates using natural language relative to the total number of users exposed to the feature in the last four weeks. 

\begin{table*}[t!]
\caption{Relative results of A/B testing compared with the baseline.}
\centering
%
\begin{tabular}[b]{llccccc}
\cline{3-7}
                                                  &             & \multicolumn{2}{c}{Latency} &                                                          &                                                         &                                                         \\ \cline{3-7} 
                                                  &             & Median        & p99         & Weekly Adoption                                           & Engagement                                              & Failures                                                \\ \hline
\multicolumn{1}{c}{\multirow{2}{*}{Experiment 1}} & Baseline    & 1.77s         & 5.31s       & 9.6\%                                                    & 1.7\%                                                   & 35\%                                                    \\
\multicolumn{1}{c}{}                              & T5QL        & 1.45s         & 5.90s       & \begin{tabular}[c]{@{}c@{}}17.0\%\\ (1.8x)\end{tabular}  & \begin{tabular}[c]{@{}c@{}}3.7\%\\ (2.2x)\end{tabular}  & \begin{tabular}[c]{@{}c@{}}6\%\\ (0.2x)\end{tabular}    \\ \hline
Experiment 2                                      & T5QL        & 0.87s         & 7.91s       & 23.5\%                                                   & 3.6\%                                                   & 4.2\%                                                   \\ 
                                                  & T5QL + FP16 & 0.92s         & 5.45s       & \begin{tabular}[c]{@{}c@{}}22.8\%\\ (0.97x)\end{tabular} & \begin{tabular}[c]{@{}c@{}}3.7\%\\ (1.03x)\end{tabular} & \begin{tabular}[c]{@{}c@{}}3.4\%\\ (0.81x)\end{tabular} \\ \hline
\end{tabular}
\label{tab:ab-testing}
\end{table*}


\subsubsection{\textbf{Engagement}} Percentage of successful aggregates of the total number of created aggregates in the last four weeks. An aggregate is successful when it is not deleted, edited, or registered with errors. 

\subsubsection{\textbf{Failures}} Percentage of invalid aggregates relative to the total number of suggested aggregates in the last four weeks. An aggregate is considered invalid when it doesn't meet all the validation criteria.


\subsubsection{\textbf{Results}}
In Table~\ref{tab:ab-testing}, we present the results for two A/B testing experiments. In the first experiment, we compare our baseline (RATSQL) against T5QL, while in the second experiment, we compare a half-precision version of T5QL, which we call T5QL + FP16, to our initial T5QL model (which is in full precision). As we can see in Table~\ref{tab:ab-testing}, the first experiment validates the considerable improvement of T5QL compared to our baseline. Although the latencies are similar (our baseline has a slightly inferior latency in the 99th percentile but a slightly bigger median latency) the number of failures considerably decreased and the business metrics (i.e., adoption and engagement) also had considerable improvements. On the other hand, on the second experiment, our control group (T5QL) and the group of users with the new feature (T5QL+FP16) don't show statistically significant differences (given the sample size collected and the differences observed) in the business metrics. We can observe that the 99th percentile of latencies decreased considerably in half-precision, but that didn't correlate to the expected improvement in the business metrics.

The roll-out plan until putting the feature into \gls{ga} was done in 3 different steps. First, we opened the feature to 40\% of the users, 15 days later was opened to 70\%, and 15 days after we released it to all OutSystems users.

\section{Related Work}\label{sec:related_work}

\quad To build the NL2Agg pipeline and put it in production with the model with the best performance possible, we explored two main concepts: hybrid-centric development and A/B testing. Our pipeline relies heavily on the data, thus allowing us to have a curated dataset to train our model and a SQL template distribution that mimics our users' usage. After this, we tested different models and train datasets. Adopting A/B testing allowed us to drastically reduce the need to test many different approaches. Finally, a crucial aspect of our work is the NL2SQL model performance. In this section, we review the three concepts and explore previous works on these research areas.


\subsubsection*{\textbf{A/B testing}}
\quad A/B testing has been used for website development since the rise of the Internet~\cite{kohavi2017online}. This technique relies on doing online-controlled experiments by providing a random set of users a control version of the website nominative A. The other set of users will be presented with version B, with some alterations to the previous version.  Previous research was done in this area to explore online A/B testing aligned with ML model deployment~\cite{baltescu2022itemsage, li2021evolving}. In most cases, the main metric to be considered is users' engagement, such as the number of clicks or pagination. Sometimes logging is also considered~\citep{xu2015infrastructure}, especially on specific parts of the website that are being used, enabling triggered analysis.

\subsubsection*{\textbf{NL2SQL}}
Code generation has been a widely explored area in Computer Science, and SQL generation, particularly, is a very active research area. Different approaches were proposed in the last years, from enhancing schema representation using BERT contextual output~\cite{he2019x} to generating code from fragments~\cite{cheung2012inferring} or even giving a database and an example of an output table~\cite{orvalho2020squares}. 
Some authors~\cite{wang2019rat} defend the need to use an intermediate representation by combing the schema entities and the words of the natural language question using relation-aware self-attention. 
SmBoP~\cite{rubin2020smbop} uses a semi-regressive decoder instead of the autoregressive decoder employed by RATSQL~\cite{wang2019rat}, reducing the number of generation steps since each decoding step can generate many subtrees. Recently, PICARD~\cite{scholak2021picard} and T5QL~\cite{arcadinho2022t5ql} constrained the generation of \glspl{llm} to generate only syntactically correct SQL. By doing this, both works reduce the number of hallucinations of these models perform, which boosts accuracy.

\section{Conclusion}\label{sec:conclusions}
This work presents an end-to-end \gls{nl2sql} pipeline, focusing on applying the best model for a specific use case and utilizing a crowdsourced trainset mimicking our customers' needs. Our pipeline also includes a feedback loop to retrain the model for specific SQL templates where the SQL generation model is not performing as well. 
The pipeline was evaluated in two settings: offline and online. Offline experiments involved a test set mimicking users' data distribution, resulting in a significantly improved model. The model was further improved by augmenting it with business rules and creating a new ~\gls{ted} metric. Online tests showed an increase of 230\% in adoption, 220\% in engagement rate, and a 90\% reduction in error rate. Future work aims to explore model shadowing by putting a new model in general availability (GA) and evaluating it offline. This will allow for the estimation of model performance without premature deployment.

\bibliography{custom}

\begin{thebibliography}{27}
\expandafter\ifx\csname natexlab\endcsname\relax\def\natexlab#1{#1}\fi

\bibitem[{Al~Alamin et~al.(2021)Al~Alamin, Malakar, Uddin, Afroz, Haider, and
  Iqbal}]{al2021empirical}
Md~Abdullah Al~Alamin, Sanjay Malakar, Gias Uddin, Sadia Afroz, Tameem~Bin
  Haider, and Anindya Iqbal. 2021.
\newblock An empirical study of developer discussions on low-code software
  development challenges.
\newblock In \emph{2021 IEEE/ACM 18th International Conference on Mining
  Software Repositories (MSR)}, pages 46--57. IEEE.

\bibitem[{Arcadinho et~al.(2022)Arcadinho, Aparício, Veiga, and
  Alegria}]{arcadinho2022t5ql}
Samuel Arcadinho, David Aparício, Hugo Veiga, and António Alegria. 2022.
\newblock T5ql: Taming language models for sql generation.
\newblock \emph{arXiv preprint arXiv:2209.10254}.

\bibitem[{Baltescu et~al.(2022)Baltescu, Chen, Pancha, Zhai, Leskovec, and
  Rosenberg}]{baltescu2022itemsage}
Paul Baltescu, Haoyu Chen, Nikil Pancha, Andrew Zhai, Jure Leskovec, and
  Charles Rosenberg. 2022.
\newblock Itemsage: Learning product embeddings for shopping recommendations at
  pinterest.
\newblock In \emph{Proceedings of the 28th ACM SIGKDD Conference on Knowledge
  Discovery and Data Mining}, pages 2703--2711.

\bibitem[{Beetz and Harrer(2022)}]{gitops2022}
Florian Beetz and Simon Harrer. 2022.
\newblock \href {https://doi.org/10.1109/MS.2021.3119106} {Gitops: The
  evolution of devops?}
\newblock \emph{IEEE Software}, 39(4):70--75.

\bibitem[{Cheung et~al.(2012)Cheung, Solar-Lezama, and
  Madden}]{cheung2012inferring}
Alvin Cheung, Armando Solar-Lezama, and Samuel Madden. 2012.
\newblock Inferring sql queries using program synthesis.
\newblock \emph{arXiv preprint arXiv:1208.2013}.

\bibitem[{{Feng} et~al.(2020){Feng}, {Guo}, {Tang}, {Duan}, {Feng}, {Gong},
  {Shou}, {Qin}, {Liu}, {Jiang}, and {Zhou}}]{2020arXiv200208155F}
Zhangyin {Feng}, Daya {Guo}, Duyu {Tang}, Nan {Duan}, Xiaocheng {Feng}, Ming
  {Gong}, Linjun {Shou}, Bing {Qin}, Ting {Liu}, Daxin {Jiang}, and Ming
  {Zhou}. 2020.
\newblock \href {http://arxiv.org/abs/2002.08155} {{CodeBERT: A Pre-Trained
  Model for Programming and Natural Languages}}.
\newblock \emph{arXiv e-prints}, page arXiv:2002.08155.

\bibitem[{Feng et~al.(2020)Feng, Guo, Tang, Duan, Feng, Gong, Shou, Qin, Liu,
  Jiang et~al.}]{feng2020codebert}
Zhangyin Feng, Daya Guo, Duyu Tang, Nan Duan, Xiaocheng Feng, Ming Gong, Linjun
  Shou, Bing Qin, Ting Liu, Daxin Jiang, et~al. 2020.
\newblock Codebert: A pre-trained model for programming and natural languages.
\newblock \emph{arXiv preprint arXiv:2002.08155}.

\bibitem[{Guo et~al.(2021)Guo, Ainslie, Uthus, Ontanon, Ni, Sung, and
  Yang}]{guo2021longt5}
Mandy Guo, Joshua Ainslie, David Uthus, Santiago Ontanon, Jianmo Ni, Yun-Hsuan
  Sung, and Yinfei Yang. 2021.
\newblock Longt5: Efficient text-to-text transformer for long sequences.
\newblock \emph{arXiv preprint arXiv:2112.07916}.

\bibitem[{He et~al.(2019)He, Mao, Chakrabarti, and Chen}]{he2019x}
Pengcheng He, Yi~Mao, Kaushik Chakrabarti, and Weizhu Chen. 2019.
\newblock X-sql: reinforce schema representation with context.
\newblock \emph{arXiv preprint arXiv:1908.08113}.

\bibitem[{Inel et~al.(2014)Inel, Khamkham, Cristea, Dumitrache, Rutjes, Ploeg,
  Romaszko, Aroyo, and Sips}]{inel2014crowdtruth}
Oana Inel, Khalid Khamkham, Tatiana Cristea, Anca Dumitrache, Arne Rutjes,
  Jelle van~der Ploeg, Lukasz Romaszko, Lora Aroyo, and Robert-Jan Sips. 2014.
\newblock Crowdtruth: Machine-human computation framework for harnessing
  disagreement in gathering annotated data.
\newblock In \emph{International semantic web conference}, pages 486--504.
  Springer.

\bibitem[{Kohavi and Longbotham(2017)}]{kohavi2017online}
Ron Kohavi and Roger Longbotham. 2017.
\newblock Online controlled experiments and a/b testing.
\newblock \emph{Encyclopedia of machine learning and data mining},
  7(8):922--929.

\bibitem[{Lewis et~al.(2019)Lewis, Liu, Goyal, Ghazvininejad, Mohamed, Levy,
  Stoyanov, and Zettlemoyer}]{lewis2019bart}
Mike Lewis, Yinhan Liu, Naman Goyal, Marjan Ghazvininejad, Abdelrahman Mohamed,
  Omer Levy, Ves Stoyanov, and Luke Zettlemoyer. 2019.
\newblock Bart: Denoising sequence-to-sequence pre-training for natural
  language generation, translation, and comprehension.
\newblock \emph{arXiv preprint arXiv:1910.13461}.

\bibitem[{Li et~al.(2021)Li, Chai, Campbell, Liao, Abburu, Kang, Niculescu,
  Brake, Patil, Dooley et~al.}]{li2021evolving}
Paul~Luo Li, Xiaoyu Chai, Frederick Campbell, Jilong Liao, Neeraja Abburu,
  Minsuk Kang, Irina Niculescu, Greg Brake, Siddharth Patil, James Dooley,
  et~al. 2021.
\newblock Evolving software to be ml-driven utilizing real-world a/b testing:
  experiences, insights, challenges.
\newblock In \emph{2021 IEEE/ACM 43rd International Conference on Software
  Engineering: Software Engineering in Practice (ICSE-SEIP)}, pages 170--179.
  IEEE.

\bibitem[{Orvalho et~al.(2020)Orvalho, Terra-Neves, Ventura, Martins, and
  Manquinho}]{orvalho2020squares}
Pedro Orvalho, Miguel Terra-Neves, Miguel Ventura, Ruben Martins, and Vasco
  Manquinho. 2020.
\newblock Squares: a sql synthesizer using query reverse engineering.
\newblock \emph{Proceedings of the VLDB Endowment}, 13(12):2853--2856.

\bibitem[{Raffel et~al.(2020)Raffel, Shazeer, Roberts, Lee, Narang, Matena,
  Zhou, Li, Liu et~al.}]{raffel2020exploring}
Colin Raffel, Noam Shazeer, Adam Roberts, Katherine Lee, Sharan Narang, Michael
  Matena, Yanqi Zhou, Wei Li, Peter~J Liu, et~al. 2020.
\newblock Exploring the limits of transfer learning with a unified text-to-text
  transformer.
\newblock \emph{J. Mach. Learn. Res.}, 21(140):1--67.

\bibitem[{Roh et~al.(2019)Roh, Heo, and Whang}]{roh2019survey}
Yuji Roh, Geon Heo, and Steven~Euijong Whang. 2019.
\newblock A survey on data collection for machine learning: a big data-ai
  integration perspective.
\newblock \emph{IEEE Transactions on Knowledge and Data Engineering},
  33(4):1328--1347.

\bibitem[{Rubin and Berant(2020)}]{rubin2020smbop}
Ohad Rubin and Jonathan Berant. 2020.
\newblock Smbop: Semi-autoregressive bottom-up semantic parsing.
\newblock \emph{arXiv preprint arXiv:2010.12412}.

\bibitem[{Sahay et~al.(2020)Sahay, Indamutsa, Di~Ruscio, and
  Pierantonio}]{sahay2020supporting}
Apurvanand Sahay, Arsene Indamutsa, Davide Di~Ruscio, and Alfonso Pierantonio.
  2020.
\newblock Supporting the understanding and comparison of low-code development
  platforms.
\newblock In \emph{2020 46th Euromicro Conference on Software Engineering and
  Advanced Applications (SEAA)}, pages 171--178. IEEE.

\bibitem[{Scholak et~al.(2021)Scholak, Schucher, and
  Bahdanau}]{scholak2021picard}
Torsten Scholak, Nathan Schucher, and Dzmitry Bahdanau. 2021.
\newblock Picard: Parsing incrementally for constrained auto-regressive
  decoding from language models.
\newblock \emph{arXiv preprint arXiv:2109.05093}.

\bibitem[{Shi et~al.(2021)Shi, Ng, Wang, Zhu, Li, Wang, dos Santos, and
  Xiang}]{shi2021learning}
Peng Shi, Patrick Ng, Zhiguo Wang, Henghui Zhu, Alexander~Hanbo Li, Jun Wang,
  Cicero~Nogueira dos Santos, and Bing Xiang. 2021.
\newblock Learning contextual representations for semantic parsing with
  generation-augmented pre-training.
\newblock In \emph{Proceedings of the AAAI Conference on Artificial
  Intelligence}, volume~35, pages 13806--13814.

\bibitem[{Wang et~al.(2019)Wang, Shin, Liu, Polozov, and
  Richardson}]{wang2019rat}
Bailin Wang, Richard Shin, Xiaodong Liu, Oleksandr Polozov, and Matthew
  Richardson. 2019.
\newblock Rat-sql: Relation-aware schema encoding and linking for text-to-sql
  parsers.
\newblock \emph{arXiv preprint arXiv:1911.04942}.

\bibitem[{Xu et~al.(2015)Xu, Chen, Fernandez, Sinno, and
  Bhasin}]{xu2015infrastructure}
Ya~Xu, Nanyu Chen, Addrian Fernandez, Omar Sinno, and Anmol Bhasin. 2015.
\newblock From infrastructure to culture: A/b testing challenges in large scale
  social networks.
\newblock In \emph{Proceedings of the 21th ACM SIGKDD International Conference
  on Knowledge Discovery and Data Mining}, pages 2227--2236.

\bibitem[{Yu et~al.(2019)Yu, Zhang, Er, Li, Xue, Pang, Lin, Tan, Shi, Li
  et~al.}]{yu2019cosql}
Tao Yu, Rui Zhang, He~Yang Er, Suyi Li, Eric Xue, Bo~Pang, Xi~Victoria Lin,
  Yi~Chern Tan, Tianze Shi, Zihan Li, et~al. 2019.
\newblock Cosql: A conversational text-to-sql challenge towards cross-domain
  natural language interfaces to databases.
\newblock \emph{arXiv preprint arXiv:1909.05378}.

\bibitem[{Yu et~al.(2018)Yu, Zhang, Yang, Yasunaga, Wang, Li, Ma, Li, Yao,
  Roman et~al.}]{yu2018spider}
Tao Yu, Rui Zhang, Kai Yang, Michihiro Yasunaga, Dongxu Wang, Zifan Li, James
  Ma, Irene Li, Qingning Yao, Shanelle Roman, et~al. 2018.
\newblock Spider: A large-scale human-labeled dataset for complex and
  cross-domain semantic parsing and text-to-sql task.
\newblock \emph{arXiv preprint arXiv:1809.08887}.

\bibitem[{Zhan et~al.(2022)Zhan, Wang, Huang, Xiong, Dou, and
  Chan}]{Zhan2022ACS}
Xueying Zhan, Qingzhong Wang, Kuan-Hao Huang, Haoyi Xiong, Dejing Dou, and
  Antoni~B. Chan. 2022.
\newblock A comparative survey of deep active learning.
\newblock \emph{ArXiv}, abs/2203.13450.

\bibitem[{Zhao et~al.(2018)Zhao, Harper, Adomavicius, and
  Konstan}]{feedback2018}
Qian Zhao, F.~Maxwell Harper, Gediminas Adomavicius, and Joseph~A. Konstan.
  2018.
\newblock \href {https://doi.org/10.1145/3167132.3167275} {Explicit or implicit
  feedback? engagement or satisfaction? a field experiment on
  machine-learning-based recommender systems}.
\newblock In \emph{Proceedings of the 33rd Annual ACM Symposium on Applied
  Computing}, SAC '18, page 1331–1340, New York, NY, USA. Association for
  Computing Machinery.

\bibitem[{Zhong et~al.(2020)Zhong, Yu, and Klein}]{zhong2020semantic}
Ruiqi Zhong, Tao Yu, and Dan Klein. 2020.
\newblock Semantic evaluation for text-to-sql with distilled test suites.
\newblock \emph{arXiv preprint arXiv:2010.02840}.

\end{thebibliography}
\bibliographystyle{acl_natbib}

\appendix

\section{Appendix}
\label{sec:appendix}

\subsection{OutSystems Aggregate}
\label{subsec:osagg}

\setcounter{figure}{0} 
\renewcommand{\thefigure}{S\arabic{figure}}

\begin{figure*}[ht]
    \centering
    \includegraphics[width=0.99\linewidth]{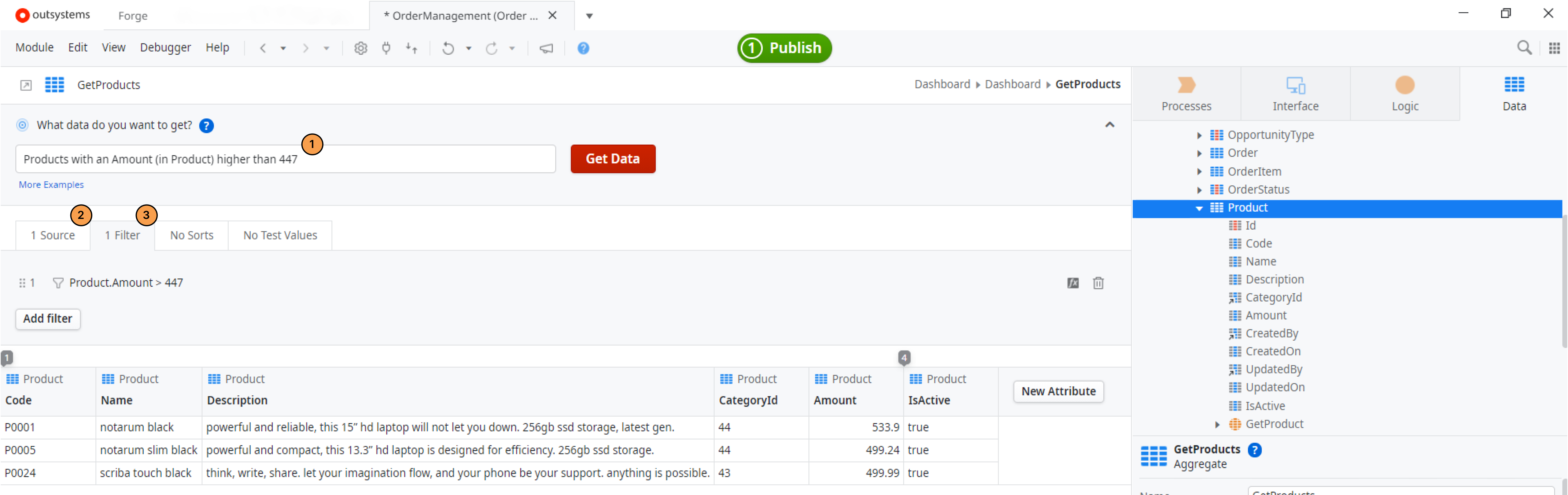}
    \caption{OutSystems IDE with the result of a natural language query. We can see the overall experience of creating an aggregate using \gls{nl} and how one can correct/change the predicted result. The normal usage of this feature starts by writing a \gls{nl} query in \circled{1}, then verify if all the entities selected are correct \circled{2} and checking possible filters \circled{3}.}
    \label{fig:aggregate}
\end{figure*}

In an aggregate, the developer can select entities \circled{2} and apply a set of filters \circled{3}. 
The result of the aggregate, like the result of a SQL query, is later shown to end-users on the screen (e.g., as a list or a table). However, low-code developers are typically not SQL experts, and thus creating aggregates can be challenging and time-consuming.

\subsection{Mturk}
\label{subsec:mturk}

\begin{figure*}[ht]
    \centering
    \includegraphics[width=0.99\linewidth]{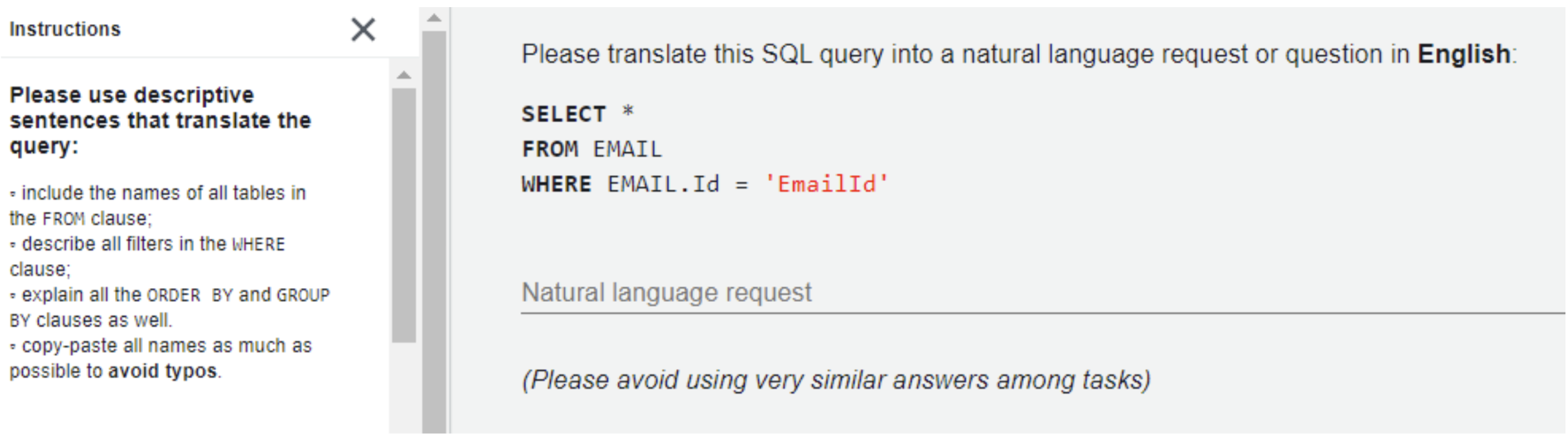}
    \caption{This image represents what the user saw when labeling a SQL query. In total, we did 12 different experiments to perfect the MTurk task.}
    \label{fig:MTurk_experiment}
\end{figure*}

The MTurk task is represented in Figure~\ref{fig:MTurk_experiment}. 

\begin{figure*}[ht]
    \centering
    \includegraphics[width=0.99\linewidth]{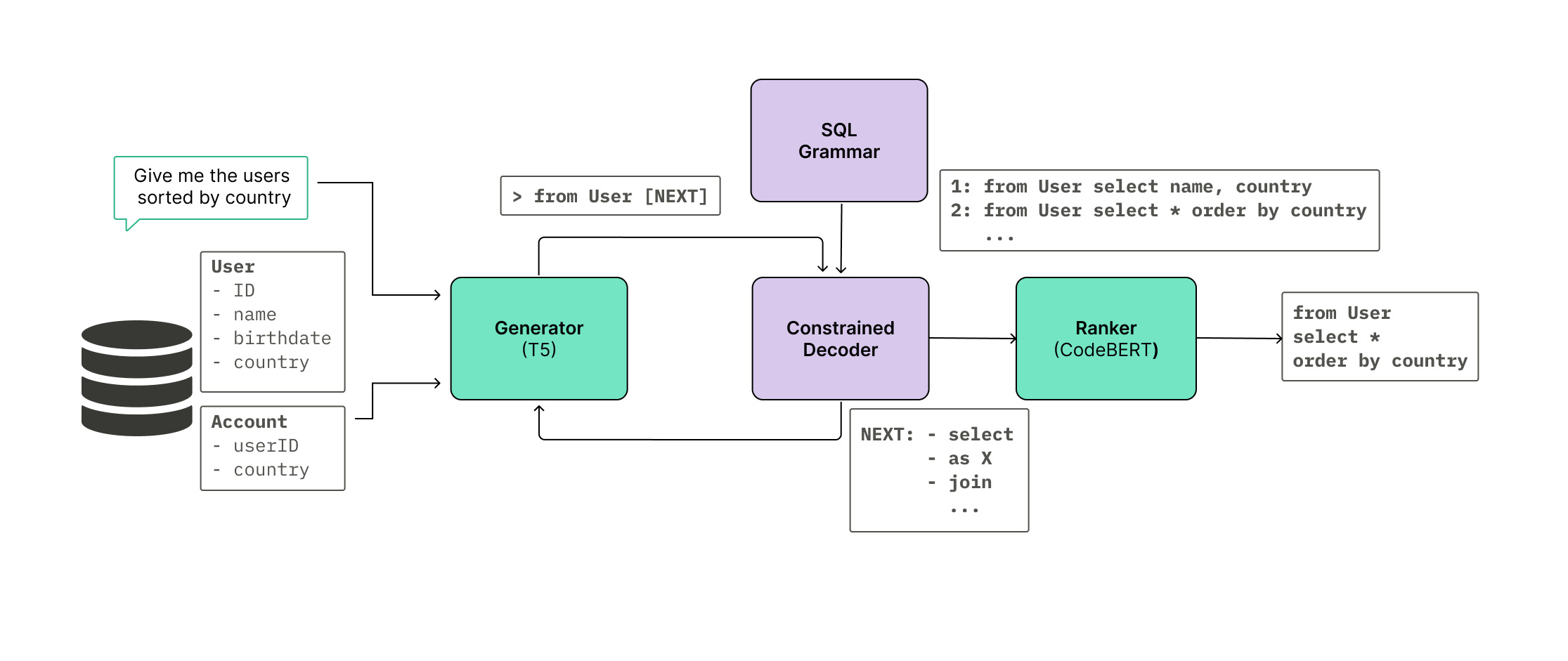}
    \caption{T5QL model architecture (from \citet{arcadinho2022t5ql}).}
    \label{fig:t5ql}
\end{figure*}

After the experiment is done we used first approach to achieve automatic validation, we used CrowdTruth~\cite{inel2014crowdtruth}. The goal of this library is to process the results from \Gls{mturk} and measure the inter-annotation agreement quality between the workers and the units (in our case, the units are the SQL and \gls{nl} pairs). The framework uses the inter-dependency between the three main components of a crowdsourcing system: worker, input data, and annotation. However, the metrics calculated by CrowdTruth did not show a correlation between the worker and the resulting \gls{nl}. Thus, we use the features provided by \Gls{mturk} and do feature engineering on the labeled pairs. The resulting features are the following: 

\begin{itemize}
    \item \textbf{Work time} - time spent by the worker (in seconds) writing the NL for the input SQL;
    \item \textbf{Has more than two words} - check if the NL has more than two words; 
    \item \textbf{Is terminal on NL} - inspect if all the terminals are present on the NL;
    \item \textbf{SQL Complexity} -  measure the difficulty of the SQL based on the number of SQL components in the query~\cite{yu2018spider}.  E.g the query SELECT COUNT(*) FROM cars\_data WHERE cylinders > 4 is classified with a hardness Easy.
    \item \textbf{Time per Complexity} - dividing the time spent on a query by its hardness;
    \item \textbf{SQL question Levenshtein distance} - calculate the Levenshtein distance between the NL and the SQL query;
    \item \textbf{Order by direction} - measure if the \gls{nl} has the most frequent words to describe the directions (ASC and DESC) of the SQL queries. A dictionary with the most frequent terms was built;
    \item \textbf{Limit Value} - similar to the previews variable. In this one, we construct a dictionary of words describing the SQL query's limit. It is a boolean that determines whether the NL has presented the words that state the limit of the SQL. 
    
\end{itemize}

\end{document}